\pdfoutput=1

\documentclass[11pt]{article}

\usepackage{acl}

\usepackage{times}
\usepackage{latexsym}

\usepackage[T1]{fontenc}

\usepackage[utf8]{inputenc}

\usepackage{microtype}

\usepackage{enumitem}
\usepackage{algorithm}
\usepackage{algorithmic}
\usepackage{amsmath}
\usepackage{multirow}
\usepackage{booktabs}
\usepackage{url}
\usepackage{graphicx}
\definecolor{RoseQuartzBg}{HTML}{F7CAC9}
\definecolor{RoseQuartz}{HTML}{F5A798}
\definecolor{Serenity}{HTML}{92A8D1}
\definecolor{OrangeRed}{rgb}{1.0, 0.27, 0.0}
\definecolor{Red}{rgb}{1.0, 0.0, 0.0}
\definecolor{Turquoise}{HTML}{0F4C81}
\usepackage{xparse}
\NewDocumentCommand{\lifu}{ mO{} }{\textcolor{OrangeRed}{\textsuperscript{\textit{Lifu}}\textsf{\textbf{\small[#1]}}}}
\NewDocumentCommand{\sijia}{ mO{} }{\textcolor{blue}{\textsuperscript{\textit{Sijia}}\textsf{\textbf{\small[#1]}}}}
\usepackage{amssymb}
\usepackage{array}
\newcolumntype{L}{>{\centering\arraybackslash}m{3cm}}
\newcolumntype{S}{>{\centering\arraybackslash}m{2cm}}
\newcolumntype{P}{>{\arraybackslash}m{10cm}}
\newcolumntype{Q}{>{\arraybackslash}m{5cm}}

\title{The Art of Prompting: Event Detection based on Type Specific Prompts}

\author{Sijia Wang$^{\clubsuit}$, \ Mo Yu$^{\spadesuit}$, \ Lifu Huang$^{\clubsuit}$
\\
  $^{\clubsuit}$Virginia Tech, \ \ 
  $^{\spadesuit}$WeChat AI
 \\
  $^{\clubsuit}${\tt \{sijiawang,lifuh\}@vt.edu}, \ \ 
  $^{\spadesuit}${\tt moyumyu@tencent.com} 
  }

\begin{document}
\maketitle
\begin{abstract}

We compare various forms of prompts to represent event types and develop a unified framework to incorporate the event type specific prompts for supervised, few-shot, and zero-shot event detection. The experimental results demonstrate that a well-defined and comprehensive event type prompt can significantly improve the performance of event detection, especially when the annotated data is scarce (few-shot event detection) or not available (zero-shot event detection). By leveraging the semantics of event types, our unified framework shows up to 24.3\% F-score gain over the previous state-of-the-art baselines.


\vspace{1cm}
\end{abstract}

\section{Introduction}
\label{sec:introduction}

Event detection~\cite{grishman1997information,chinchor1998muc,ahn2006stages} is the task of identifying and typing event mentions from natural language text. Supervised approaches, especially deep neural networks~\cite{chen-etal-2020-reading, xinyaduEMNLP2020,yinglinACL2020, jianliu2020emnlp, EEMQA_li, Lyu-etal-2021-zero}, have shown remarkable performance under a critical prerequisite of a large amount of manual annotations. However, they cannot be effectively generalized to new languages, domains or types, especially when the annotations are not enough~\cite{huang2016liberal,huang2020semi,extensively_lai_2020, Adaptive_Shirong_2021} or there is no annotations available \cite{Lyu-etal-2021-zero, zhang-etal-2021-zero, pasupat2014zero}.

\begin{table}[h]
\centering
\small
\begin{tabular}{S|Q}
\toprule
    Type Name & Attack\\
\midrule
    \multirow{2}{*}{Definition}      & Violent or physical act causing harm \\& or damage\\
\midrule
    Seed Trigger    & Invaded, airstrikes, overthrew, ambushed\\
\midrule
    Type Structure       & Attack, Attacker, Instrument, Victim, Target, Place\\
\midrule
   APEX Prompt       &  Attack, invaded airstrikes overthrew ambushed, an Attacker physically attacks a Target with Instrument at a Place \\
\bottomrule
\end{tabular}
\caption{Example of various forms prompt for the event type \textit{Conflict: Attack}}
\label{tab:attack_example}
\end{table}
Recent studies have shown that both the accuracy and generalizability of event detection can be improved via leveraging the semantics of event types based on various forms of prompts, such as event type specific queries~\cite{Lyu-etal-2021-zero, xinyaduEMNLP2020, jianliu2020emnlp}, definitions~\cite{chen-etal-2020-reading}, structures~\cite{yinglinACL2020, 10.1145/3308558.3313562}, or a few prototype event triggers~\cite{ wang2009character,Dalvi2012WebSetsES,pasupat2014zero,bronstein2015seed, lai2019extending, zhang-etal-2021-zero, Cong2021FewShotED}.
Table \ref{tab:attack_example} shows an example of each form of event type prompt for detecting event mentions from the input sentence. These studies further encourage us to take another step forward and think about the following three questions: (1) does the choice of prompt matter when the training data is abundant or scarce? (2) what's the best form of prompt for event detection? (3) how to best leverage the prompt to detect event mentions? 

To answer the above research questions, we conduct extensive experiments with various forms of prompts for each event type, including (a) \textit{event type name}, (b) \textit{prototype seed triggers}, (c) \textit{definition}, (d) \textit{event type structure} based on both event type name and its predefined argument roles, (e) free parameter based \textit{continuous soft prompt}, and (f) a more comprehensive event type description (named \textit{APEX prompt}) that covers all the information of prompts (a)-(d), under the settings of supervised event detection, few-shot and zero-shot event detection. We observe that (1) by considering the semantics of event types with most forms of prompts, especially seed triggers and the comprehensive event type descriptions, the performance of event detection under all settings can be significantly improved; (2) Among all forms of event representations, the comprehensive description based prompts show to be the most effective, especially for few-shot and zero-shot event detection; (3) Different forms of event type representations provide complementary improvements, indicating that they capture distinct aspects and knowledge of the event types.

In summary, our work makes the following contributions: 
\begin{itemize}[noitemsep,nolistsep,wide]
\item{we investigate various forms of prompts to represent event types for both supervised and weakly supervised event detection, and prove that a well-defined and comprehensive event type prompt can dramatically improve the performance of event detection and the transferability from old types to new types.}
\item{we developed a unified framework to leverage the semantics of event types with prompts for supervised, few-shot and zero-shot event detection, and demonstrate state-of-the-art performance with up to 24.3\% F-score improvement over the strong baseline methods.}
\end{itemize}

\section{Related Work}

\label{sec:related_work}




\paragraph{Supervised Event Detection:}
Most of the existing Event Detection studies follow a supervised learning paradigm~\cite{ji2008refining,liao2010using,mcclosky2011event,qiliACl2013,chen2015event,cao-etal-2015-improving,feng-etal-2016-language,yang-mitchell-2016-joint,nguyen_jrnn_2016,zhang2017improving,yinglinACL2020,cleve}, however, they cannot be directly applied to detect new types of events. Recently studies have shown that, by leveraging the semantics of event types based on type-specific questions~\cite{xinyaduEMNLP2020, jianliu2020emnlp, EEMQA_li, Lyu-etal-2021-zero} or seed event triggers~\cite{bronstein2015seed,lai2019extending, wang2021query}, the event detection performance can be improved. However, it's still unknown that whether they are the best choices of representing the semantics of event types.

\paragraph{Few-shot Event Detection:} Two primary learning strategies in few-shot classification tasks are Meta-Learning~\cite{Kang_ICCV_2019, Li2021BeyondMC, Xiao2020FewShotOD, Yan2019MetaRT, Chowdhury2021FewshotIC}, and Metric Learning~\cite{Sun2021FSCEFO, wang2020few, Zhang2021PNPDetEF, agarwal2021attention}. Several studies have exploited metric learning to align the semantics of candidate events with few examples of the novel event types for few-shot event detection~\cite{Exploiting_Lai_2020, Meta_Deng_2020, extensively_lai_2020, Cong2021FewShotED, honey_chen_2021,Adaptive_Shirong_2021}. However, due to the limited annotated data and the diverse semantics of event mentions, it's hard to design a metric distance to accurately capture the semantic similarity between the seed mentions and new ones.

\paragraph{Zero-shot Event Detection:} The core idea of zero-shot learning is to learn a mapping function between seen classes and their corresponding samples, and then apply it to ground new samples to unseen classes.~\newcite{huang2017zero} first exploited zero-shot event extraction by leveraging Abstract Meaning Representation~\cite{banarescu-etal-2013-amr} to represent event mentions and types into a shared semantic space. Recent studies~\cite{zhang-etal-2021-zero,Lyu-etal-2021-zero} further demonstrate that without using any training data, by leveraging large external corpus with abundant anchor triggers, zero-shot event detection can also be achieved with decent performance. However, such approaches cannot properly identify event mentions, i.e., distinguishing event mentions from none-event tokens.

\paragraph{Prompt Learning} Prompt learning aims to learn a task-specific prompt while keeping most of the parameters of the model freezed~\cite{Li2021PrefixTuningOC, hambardzumyan-etal-2021-warp, gpt3}. It has shown competitive performance in a wide variety of applications in natural language processing~\cite{Raffel2020ExploringTL, gpt3, autoprompt_emnlp20, Jiang2020HowCW, Lester2021ThePO, Schick2021naacl}. Previous work either use a manual \cite{petroni-etal-2019-language, gpt3, Schick_emnlp2021_few} or automated approach \cite{Jiang2020HowCW,yuan2021bartscore,Li2021PrefixTuningOC} to create prompts. In this work, we compare various forms of template based and free-parameter based prompts for event detection task under both supervised and weakly supervised setting.
\begin{figure*}
  \centering
  \includegraphics[width=0.85\textwidth]{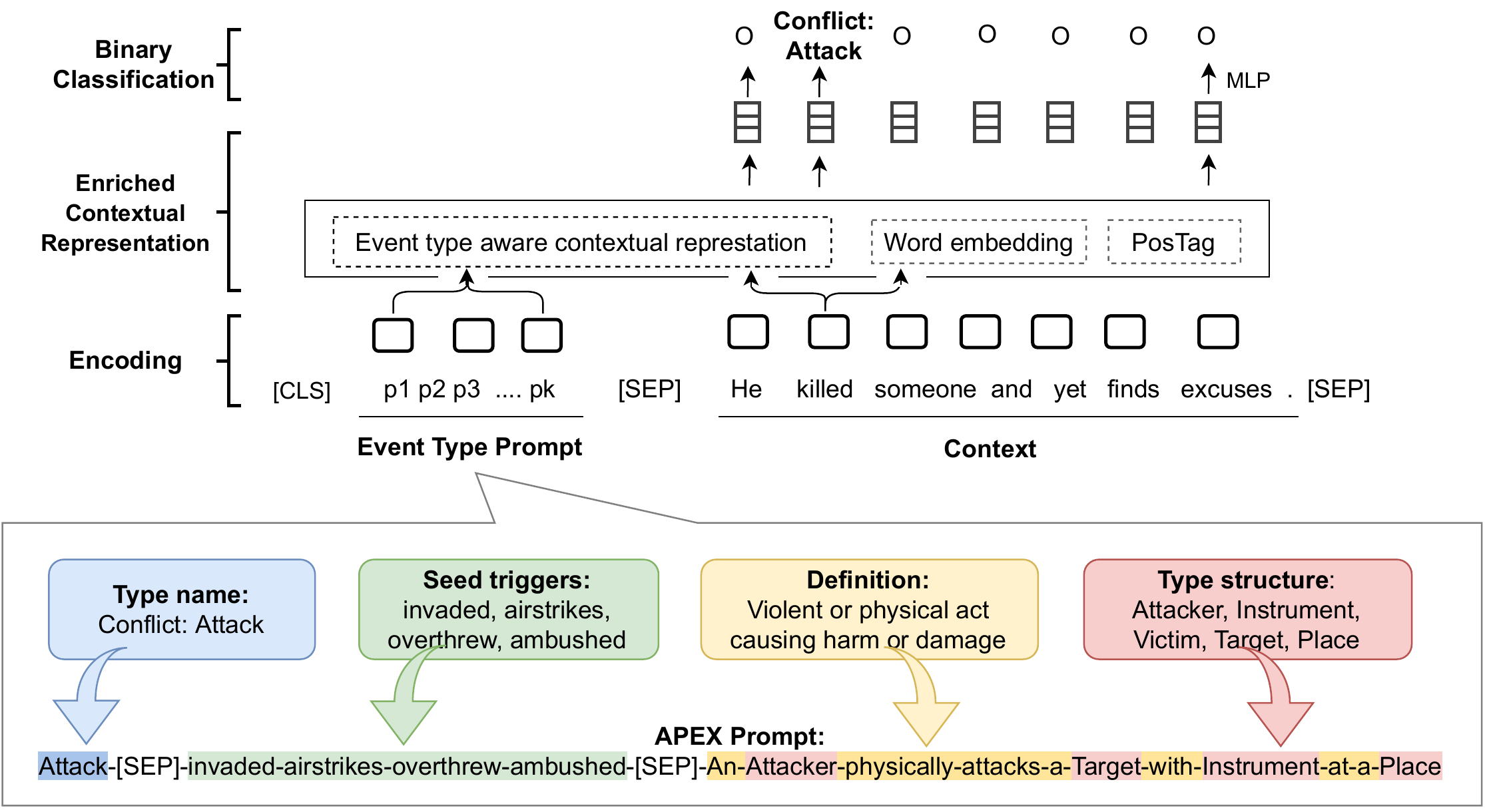}
  \caption{Overview of the unified framework for event detection based on event type specific prompts.}
  \label{fig:architecture}
\end{figure*}


\section{Problem Formulation}
\label{sec:model}
In this work, we aim to compare various forms of prompts to represent the event types under different settings, including supervised event detection, few-shot event detection and zero-shot event detection. Here, we first provide a definition for each setting of the event detection task and then describe the various forms of event type prompts.

\subsection{Settings of Event Detection}

\paragraph{Supervised Event Detection} We follow the conventional supervised event detection setting where both the training, validation and evaluation data sets cover the same set of event types. The goal is to learn a model $f$ on the training data set and evaluate its capability on correctly identifying and classifying event mentions for the target event types.

\paragraph{Few-shot Event Detection} There are two separate training data sets for few-shot event detection: (1) A large-scale base training data set $\mathcal{D}_{base}=\{(\mathbf{x}_i, \mathbf{y}_i)\}_{i=1}^{M}$ that covers the old event types (named \textit{base types}) with abundant annotations and $M$ denotes the number of base event types; (2) a smaller training data set $\mathcal{D}_{novel}=\{(\mathbf{x}_j, \mathbf{y}_j)\}_{j=1}^{N\times K}$ that covers $N$ novel event types, with $K$ examples each. Note that the base and novel event types are disjoint except the \texttt{Other} class. The model $f$ will be first optimized on $\mathcal{D}_{base}$, and then further fine-tuned on $D_{novel}$. The validation data set contains the mentions of both base and novel event types, while the evaluation data set only includes mentions of novel event types. The goal is to evaluate the generalizability and transferability of the model from base event types to new event types with few annotations.

\paragraph{Zero-shot Event Detection} The only difference between zero-shot and few-shot event detection lies in the training data sets. In zero-shot event detection, there is only a large-scale base training data set $\mathcal{D}_{base}=\{(\mathbf{x}_i, \mathbf{y}_i)\}_{i=1}^{M}$ with sufficient annotations for the base event types. The model $f$ will be only optimized on base event types and evaluated on the novel types, which is to measure the transferability of the model under a more challenging setting.

\subsection{Event Type Prompts}
We compare the following five forms of prompts to represent the event types:  


\paragraph{Event Type Name} The most straightforward and intuitive representation of an event type is the type name, which usually consists of one to three tokens. As the most basic and discriminative representations of event types, we include them in all the following text-based event type prompts.


\paragraph{Definition} The type name sometimes cannot accurately represent the semantics of an event type due to the ambiguity of the type name as well as the variety of the event mentions. For example, \textit{execute} can either refer to \textit{putting a legal punishment into action} or \textit{performing a skillful action or movement}. 
The definitions instead formally describe the meaning of the event types. Taking the event type \textit{Attack} from ACE as an example, its definition is \textit{violent or physical act causing harm or damage}

\paragraph{Prototype Seed Triggers} Seed trigger based representation consists of the type name and a list of prototype triggers. Given an event type $t$ and its annotated triggers, following~\cite{wang2021query}, we select the top-$K$\footnote{In our experiments, we set $K=4$.} ranked words as the prototype triggers based on the probability $f_t/f_o$ of each word, where $f_o$ is the frequency of the word from the whole training dataset and $f_t$ is the frequency of the word being tagged as an event trigger of type $t$. Thus, for the event type \textit{Attack}, we represent it as \textit{attack invaded airstrikes overthrew ambushed}. 



\paragraph{Event Type Structure} Each event is associated with several arguments, indicating the core participants. Our preliminary experiment shows that for certain event types, the arguments can help determine the existence of its corresponding events. For example, given a sentence, if no person presents in the context, there should be no \textit{Meet} events. Given that, we define an event type structure, which consists of the event type name and argument roles, to represent the event type, e.g., \textit{attack attacker victim target instrument place} for \textit{Attack}.


\paragraph{Continuous Soft Prompt} Inspired by the recent success of prompt tuning methods in various NLP applications, we also adopt a continuous soft prompt, i.e., a free vector of parameter, to represent each event type. More details regarding the learning of soft prompts are described in Section~\ref{sec:unified_ed}.

\paragraph{APEX Prompt} We assume a better representation of an event type should cover the important information of all the above prompts. Thus we define a more comprehensive description (named \textit{APEX prompt}) for each event type by concatenating its event type name, seed triggers, and definition which covers all the argument roles. For example, The APEX prompt for \textit{Attack} event type is \textit{attack, invaded airstrikes overthrew ambushed, an attacker physically attacks a target with an instrument at a place}.

In our experiments, the event type names and event type structures are automatically extracted from the target event ontology, such as ACE~\cite{ldc_ace05}, ERE~\cite{song2015light} and MAVEN~\cite{MAVEN}. The prototype seed triggers for each event type are automatically selected from its annotated data. The definitions and APEX prompts are based on the official annotation guides for each target event ontology~\cite{ldc_ace05,song2015light,MAVEN} and the available definitions in FrameNet~\cite{baker1998berkeley} with manual editing.



\section{A Unified Framework for Event Detection}
\label{sec:unified_ed}


Figure~\ref{fig:architecture} shows the overview of our unified framework, which leverages event type specific prompts to detect events under supervised, few-shot and zero-shot settings. Next, we will describe the details of this framework.

\paragraph{Context Encoding} Given an input sentence $W = \{w_{1}, w_{2}, \dots, w_{N} \}$, we take each event type prompt $T= \{\tau_1^t, \tau_2^t, \dots, \tau_K^t\}$ as a query to extract the corresponding event triggers. Specifically, we first concatenate them into a sequence as follows:
\begin{align}
\small
\text{[CLS]} \; \tau_1^t\; ...\; \tau_K^t\;  \text{[SEP]} \; w_{1}\; ...\; w_{N}\; \text{[SEP]} \nonumber
\end{align}
where [SEP] is a separator from the BERT encoder~\cite{devlin-etal-2019-bert}. 
We use a pre-trained BERT encoder to encode the whole sequence and get contextual representations for the input sentence $\boldsymbol{W} = \{\boldsymbol{w}_0, \boldsymbol{w}_2, ..., \boldsymbol{w}_N\}$ as well as the event type prompt $ \boldsymbol{T} = \{\boldsymbol{\tau}_0^t, \boldsymbol{\tau}_1^t, ..., \boldsymbol{\tau}_K^t\}$.\footnote{We use bold symbols to denote vectors.}

\paragraph{Event Type Aware Contextual Representation} Given a prompt of each event type, we aim to  extract corresponding event triggers from the input sentence automatically. To achieve this goal, we need to capture the semantic correlation of each input token to the event type. Thus we apply attention mechanism to learn a weight distribution over the sequence of contextual representations of the event type query for each token:
\begin{align}
\small
    & \boldsymbol{A}_{i}^{T} = \sum_{j=1}^{|T|}\alpha_{ij}\cdot\boldsymbol{T}_j,\,\, \textrm{where}\,\,\alpha_{ij} = \cos(\boldsymbol{w}_i,\ \boldsymbol{T}_j),\nonumber
\end{align}
where $\boldsymbol{T}_j$ is the contextual representation of the $j$-th token in the sequence $T= \{t, \tau_1^t, \tau_2^t, \dots, \tau_K^t\}$. $\cos(\cdot)$ is the cosine similarity function between two vectors. $\boldsymbol{A}_{i}^{T}$ denotes the event type $t$ aware contextual representation of token $w_i$.

\paragraph{Event Detection} With the aforementioned event type prompt attention, each token $w_i$ from the input sentence will obtain a enriched contextual representations $\boldsymbol{A}_{i}^{T}$. We concatenate them with the original contextual representation $\boldsymbol{w}_i$ from the encoder, and classify it into a binary label, indicating it as a candidate trigger of event type $t$ or not:
\begin{align*}
\small
  \boldsymbol{\tilde{y}}_i^t = \boldsymbol{U}_{o}([\boldsymbol{w}_i;\  \boldsymbol{A}_{i}^{T}; \boldsymbol{P}_i]) \;,
\end{align*}
where $[;]$ denotes concatenation operation, $\boldsymbol{U}_{o}$ is a learnable parameter matrix for event trigger detection, and $\boldsymbol{P}_i$ is the one-hot part-of-speech (POS) encoding of word $w_i$. 

For continuous soft prompt based event detection, we follow~\newcite{Li2021PrefixTuningOC} where a prefix index $q$ is prepended to the input sequence $W' = [q;\; W]$. The prefix embedding is learned by $\boldsymbol q=\text{MLP}_\theta(\boldsymbol Q_\theta[q])$, where $\boldsymbol Q_\theta\in \mathbb{R}^{|\mathcal Q|\times k}$ denotes the embedding lookup table for the vocabulary of prefix indices. Both $\text{MLP}_\theta$ and $\boldsymbol Q_\theta$ are trainable parameters. After obtaining the prefix embedding $\boldsymbol q$, we concatenate it with the initialized token embeddings of the input sentence and feed them to BERT encoder. For each token $w_i$, we obtain its contextual representation $\boldsymbol w_i$, concatenate it with its POS tag encoding $\boldsymbol{P}_i$, and then classify the token into a binary label.

\begin{table*}[h]
\centering
\small
\begin{tabular}{c|c|c|c|c|Q}
\toprule
\multicolumn{2}{c|}{Dataset} &  ACE05-E+ & ERE-EN & MAVEN & \multicolumn{1}{c}{Notes}
 \\
\midrule
\multirow{2}{*}{\# Types}&
{Base}  &  18 & 25    &120& - \\
&{Novel}  &  10 &  10   &45& - \\
\midrule
\multirow{2}{*}{\# Mentions}&
{ Base}  &  3572 & 5449    & 93675 & - \\
&{Novel} &  1724 & 3183    & 3201  & - \\
\midrule
\multirow{2}{*}{Train}
& Few-shot & 3216 & 3886  &88085 &    Include mentions of base types and a small set of mentions for novel types \\
& Zero-shot & 3116  & 3786  & 87635 &   Include mentions of base types \\
\midrule
\multicolumn{2}{c|}{\multirow{2}{*}{Validation} }    & 900   & 2797   &3883&  Mentions of base and novel types \\
\multicolumn{2}{c|}{}                         & ( 51\%/49\% ) & ( 53\%/47\% ) & ( 71\%/23\% )& Indicate the base/novel mention ratio \\
\midrule
\multicolumn{2}{c|}{Evaluation}       & 1195        &  2012        &1652 & Include mentions of novel types\\
\bottomrule
\end{tabular}
\caption{Data statistics for ACE2005, ERE and MAVEN datasets under the few-shot and zero-shot event detection settings.}
\label{tab:data statistics}
\end{table*}

\paragraph{Learning Strategy}
The learning strategy varies for supervised learning, few-shot learning and zero-shot learning. For supervised learning, we optimize the following objective for event trigger detection
\begin{align*}
\small
    \mathcal{L} = -\frac{1}{|\mathcal{T}||\mathcal{N}|}
        \sum_{t\in\mathcal{T}}
        \sum_{i=1}^{|\mathcal{N}|} \boldsymbol{y}_i^t\cdot\log \boldsymbol{\tilde{y}}_i^t\;,
\end{align*}
where $\mathcal{T}$ is the set of target event types and $\mathcal{N}$ is the set of tokens from the training dataset. $\boldsymbol{y}_i^t$ denotes the groundtruth label vector.

For few-shot event detection, we optimize the model on both base training data set and the smaller training data set for novel event types:
\begin{align*}
    \small\mathcal{L} = &\small -\frac{1}{|\mathcal{T}^B||\mathcal{N}^B|}
        \sum_{t\in\mathcal{T}^B}
        \sum_{i=1}^{|\mathcal{N}^B|} \boldsymbol{y}_i^t\cdot\log \boldsymbol{\tilde{y}}_i^t\\
        & \small- \alpha\frac{1}{|\mathcal{T}^N||\mathcal{N}^N|}
        \sum_{t\in\mathcal{T}^N}
        \sum_{i=1}^{|\mathcal{N}^N|} \boldsymbol{y}_i^t\cdot\log \boldsymbol{\tilde{y}}_i^t
\end{align*}
where $\mathcal{T}^B$ and $\mathcal{N}^B$ denote the set of base event types and tokens from the base training data set, respectively. $\mathcal{T}^N$ is the set of novel event types. $\mathcal{N}^N$ is the set of tokens from the training data set for novel event types. $\alpha$ is a hyper-parameter to balance the two objectives.

For zero-shot event detection, as we only have the base training data set, we minimize the following objective: 
\begin{align*}
\small
    \mathcal{L} = -\frac{1}{|\mathcal{T}^B||\mathcal{N}^B|}
        \sum_{t\in\mathcal{T}^B}
        \sum_{i=1}^{|\mathcal{N}^B|} \boldsymbol{y}_i^t\cdot\log \boldsymbol{\tilde{y}}_i^t\;.
\end{align*}

\begin{table*}[h]
\centering
\small
\begin{tabular}{l|c|c|c}
\toprule
    \textbf{Method} &  Supervised ED & Few-shot ED & Zero-shot ED\\ 
\midrule
\multirow{2}{*}{State of the art} & 73.3 & 35.2$^*$ & $ 49.1^*$\\
& \cite{vannguyen2021crosstask} 
& \cite{extensively_lai_2020}
& \cite{zhang-etal-2021-zero} \\
\midrule
(a) Event Type name & 72.2  & 52.7  & 49.8\\
(b) Definition &73.1        & 46.7  & 45.5\\
(c) Seed Triggers &73.7	    & 53.8  & 52.4\\
(d) Event Type Structure &72.8 &50.4&48.0\\
(e) Continuous Soft Prompt & 68.1 & 48.2 & - \\
\midrule
Majority Voting of (a)-(e)  & 73.9 & 52.1 & 48.7\\
\midrule
(f) \textbf{APEX Prompt} &\textbf{ 74.9 }& \textbf{57.4} &\textbf{55.3} \\
\bottomrule
\end{tabular}
\caption{Results of event detection (ED) on ACE05 (F1-score, \%) $^*$ indicates evaluation on our data set split. }
\label{tab:ACE}
\end{table*}

\begin{table*}[h]
\centering
\small
\begin{tabular}{l|c|c|c}
\toprule
    {\textbf{Method}} &  Supervised ED  & Few-shot ED & Zero-shot ED \\ 
\midrule
\multirow{2}{*}{State of the art} & 59.4 & 33.0$^*$ &  $41.2^*$\\
& \cite{text2event} 
& \cite{extensively_lai_2020}
& \cite{zhang-etal-2021-zero} \\
\midrule
    (a) Event Type Name & 58.2  & 44.8 & 40.5  \\
    (b) Definition      & 57.9  & 44.2 & 40.4  \\
    (c) Seed Triggers    & 60.4  & 50.4 & 49.8  \\
    (d) Event Type Structure       & 59.1  & 48.5 & 48.7   \\
    (e) Continuous Soft Prompt & 55.6& 41.7 & - \\
\midrule
    Majority Voting of (a)-(e)  &  60.2 & 47.9 & 48.3 \\
\midrule
    (f) \textbf{APEX Prompt} &  \textbf{63.4}  & \textbf{52.6} & \textbf{49.9} \\
\bottomrule
\end{tabular}
\caption{Results of event detection (ED) on ERE (F1-score, \%). $^*$ indicates evaluation on our data set split.}
\label{tab:ERE}
\end{table*}

\begin{table*}[h]
\centering
\small
\begin{tabular}{l|c|c|c}
\toprule
    {\textbf{Method}} &  Supervised & Few-shot & Zero-shot \\ 
\midrule
\multirow{2}{*}{State of the art} & 68.5 & 57.0 & 40.2* \\
& \cite{cleve} & \cite{honey_chen_2021}& \cite{zhang-etal-2021-zero}\\
\midrule
    (a) Event type name & 68.8 & 63.4  & 58.8 \\
    (b) Definition & 67.1 &  56.9 & 52.9\\
    (c) Seed Triggers & 68.7 & 65.1  & 62.2\\
    (e) Continuous Soft Prompt & 64.5 & 38.6 & - \\
\midrule
Majority Voting of (a)-(e) & 68.4 & 63.4 & 58.6 \\
\midrule
    (f) \textbf{APEX Prompt} & \textbf{68.8}  & \textbf{ 68.4} & \textbf{64.5} \\
\bottomrule
\end{tabular}
\caption{Results of event detection (ED) on MAVEN (F1-score, \%).  $^*$ indicates evaluation on our data set split.}
\vspace{-1.2em}
\label{tab:MAVEN}
\end{table*}

\section{Experiment Setup}
\label{sec:experiments}
\subsection{Datasets}

We perform experiments on three public benchmark datasets, include ACE05-E$^+$ (Automatic Content Extraction)\footnote{\url{https://catalog.ldc.upenn.edu/LDC2006T06}}, ERE (Entity Relation Event)~\cite{song2015light}\footnote{Following~\citet{yinglinACL2020}, we merge LDC2015E29, LDC2015E68, and LDC2015E78 as the ERE dataset.}, and MAVEN\cite{MAVEN}. On each dataset, we conduct experiments under three settings: supervised event detection, few-shot and zero-shot event detection. 

For supervised event detection, we use the same data split as the previous studies~\cite{qiliACl2013,wadden-etal-2019-entity,yinglinACL2020,xinyaduEMNLP2020,yinglinACL2020,vannguyen2021crosstask, MAVEN} on all the three benchmark datasets.

For few-shot and zero-shot event detection on MAVEN, we follow the previous study~\cite{honey_chen_2021} and choose 120 event types with the most frequent mentions as the base event types and the rest 45 event types as novel ones. 
For few-shot and zero-shot event detection on ACE and ERE, previous studies~\cite{extensively_lai_2020, Exploiting_Lai_2020, honey_chen_2021} follow different data splits and settings, making it hard for fair comparison. Considering the research goals of few-shot and zero-shot event detection, we define the following conditions to split the ACE and ERE datasets:
\begin{itemize}
    \item The base event types and novel event types should be disjoint except \texttt{Other}.
    \item Each base or novel event type should contain at least 15 instances.
    \item The training set should contain sufficient annotated event mentions.
\end{itemize}

To meet the above conditions, for ACE, we define the event types of 5 main event categories: \textit{Business}, \textit{Contact}, \textit{Conflict}, \textit{Justice} and \textit{Movement} as the base event types, and types of the remaining 3 main categories: \textit{Life}, \textit{Personnel} and \textit{Transaction} as the novel event types. In total, there are 18 qualified base types and 10 qualified novel types (the others do not satisfy the second condition). For ERE, we use the exact same 10 novel event types as ACE, and the rest 25 types as base event types. 

After defining the base and novel event types, we further create the training, validation and evaluation splits for all three datasets. For few-shot event detection, we use the sentences with only base event type mentions as the base training data set, and randomly select 10 sentences with novel event type mentions as the additional smaller training data set. We use the sentences with both base and novel event type mentions as the development set, and use the remaining sentences with only novel event type mentions as the evaluation dataset. For zero-shot event detection, we use the same development and evaluation set as few-shot event detection, and remove the instances with novel event mentions from the training set. For both zero-shot and few-shot event detection, we randomly split the sentences without any event annotations proportionally to the number of sentences with event mentions in each set. Table \ref{tab:data statistics} shows the detailed data statistics for all the three datasets under the few-shot and zero-shot event extraction settings.

\subsection{Hyperparameters and Evaluation}
For a fair comparison with the previous baseline approaches, we use the same pre-trained \texttt{bert-large-uncased} model for fine-tuning and optimizing our model with BertAdam. For supervised event detection, we optimize the parameters with grid search: training epoch 3, learning rate $\in [3e\text{-}6, 1e\text{-}4]$, training batch size $\in\{8, 12, 16, 24, 32\}$, dropout rate $\in\{0.4, 0.5, 0.6\}$. The running time is up to 3 hours on one Quadro RTX 8000. For evaluation, we use the same criteria as previous studies~\cite{qiliACl2013,chen2015event,nguyen_jrnn_2016,yinglinACL2020}: an event mention is correct if its span and event type matches a reference event mention. 

\begin{figure*}
  \centering
  \includegraphics[width=0.8\textwidth]{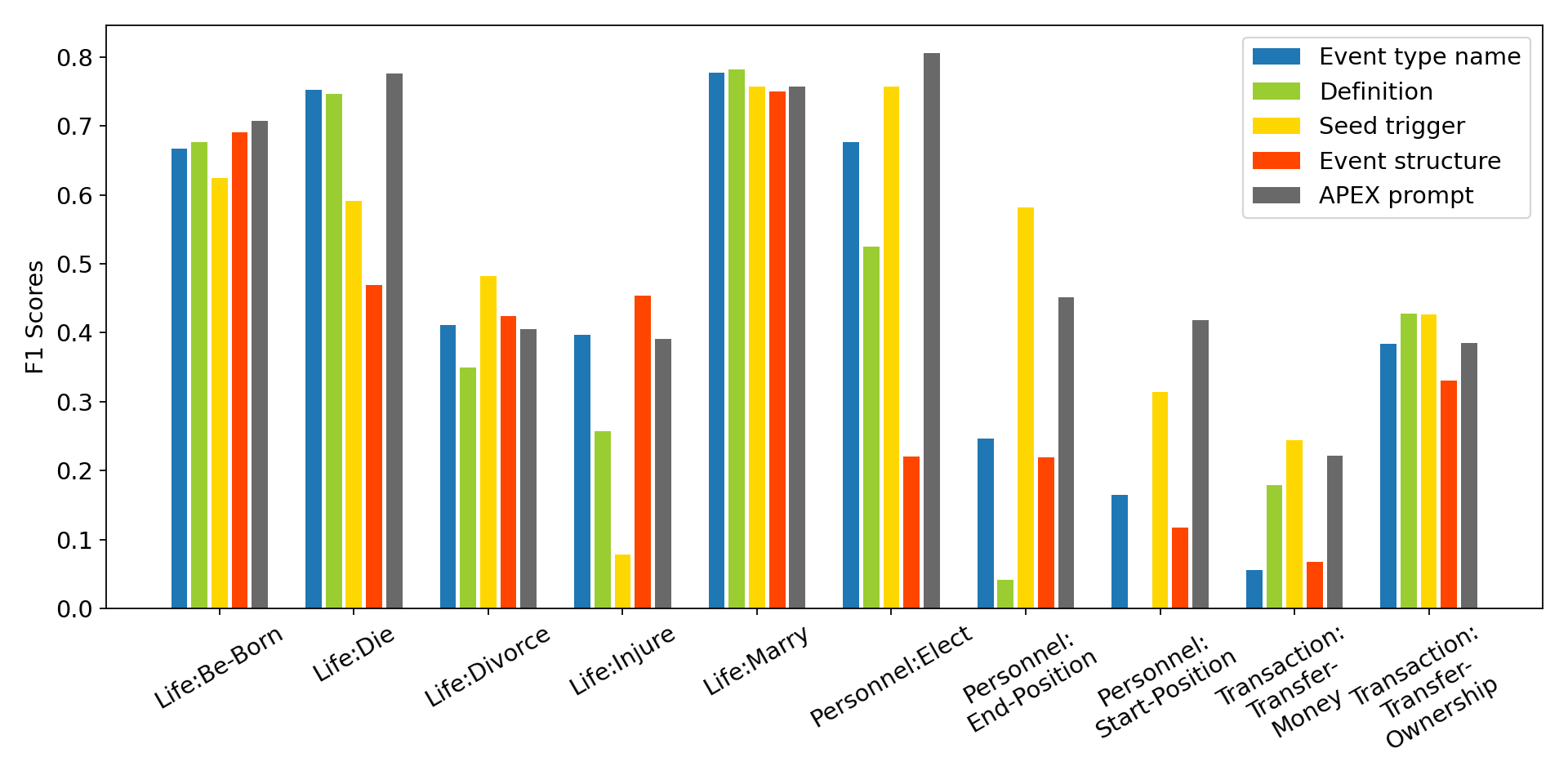}
  \vspace{-0.6em}
  \caption{F-score distribution of all novel types based on various event type prompts under the few-shot event detection setting on ACE (Best view in color)}
  \label{fig:f1_distribution}
  \vspace{-1.2em}
\end{figure*}

\section{Results and Discussion}

\paragraph{Overall Results}
The experimental results for supervised, few-shot and zero-shot event detection on ACE05, ERE and MAVEN are shown in Table~\ref{tab:ACE}-\ref{tab:MAVEN}, from which we see that (1) the APEX prompt achieves the best performance among all the forms of prompts under all the settings of the three benchmark datasets. Comparing with the previous state of the art, the APEX prompt shows up to 4\% F-score gain for supervised event detection (on ERE), 22.2\% F-score gain for few-shot event detection (on ACE), and 24.3\% F-score gain for zero-shot event detection (on MAVEN); (2) All the forms of prompts provide significant improvement for few-shot and zero-shot event detection, demonstrating the benefit of leveraging the semantics of event types via various forms of prompts for event detection, especially when the annotations are limited or not available. (3) Continuous soft prompt does not provide comparable performance as other forms of event type representations, which proves the necessity of leveraging event type specific prior knowledge to the representations; (4) The majority voting does not show improvement over individual prompts, due to the fact that each individual prompt captures a particular aspect of the event type semantics.

\paragraph{Supervised Event Detection}
By carefully investigating the event mentions that are correctly detected by the APEX prompt while missed by other prompts, we find that the APEX prompt is more effective in detecting two types of event mentions: homonyms (multiple-meaning words) and intricate words. General homonyms are usually hard to be detected as event mentions as they usually have dozens of meanings in different contexts. For example, consider the following two examples: (i) \textit{Airlines are getting [Transport:Movement] flyers to destinations on time more often .} (ii) \textit{If the board cannot vote to give [Transaction:Transfer-Money'] themselves present money}. Here, ``get'', and ``give'' are not detected based on the event type name or seed triggers but correctly identified by the definition and APEX prompts. In general, the definition and APEX prompts make 10\% and 7\% fewer false predictions than seed triggers on general homonyms. For intricate words, their semantics usually cannot be captured with an individual prompt. In the following two examples: (i) \textit{It is reasonable, however, to reimburse board members for legitimate expenses}
(ii) \textit{$\cdot\cdot\cdot$ ever having discussed being compensated by the board in the future $\cdot\cdot\cdot$}, ``reimburse'' and ``compensated'' indicate sophisticated meaning of \textit{Transaction:Transfer-Money}, which may not be captured by prompts, such as seed triggers. With the event definition and the argument roles in the APEX prompt, the highly correlated contexts, such as ``board members'' and ``legitimate expenses'', can help the model correctly detect \textit{reimburse} as an event mention of \textit{Transaction:Transfer-Money}.

\paragraph{Few-shot Event Detection}
Figure~\ref{fig:f1_distribution} shows the F-score distribution of all novel types based on various forms of event type prompts, from which we observe that:
(1) The event type name, seed triggers, and APEX prompt generally perform better than definition and structure, as they carry more straightforward semantics of event types. (2) Event type name based prompts show lower performance on \textit{Personnel:End-Position}, \textit{Personnel:Start-Position} and \textit{Transaction:Transfer-Money} than other event types, as the semantics of these event type names are less indicative than other event types. (3) Seed triggers based prompts perform worse than event type name and APEX prompts on two event types, \textit{Life:injure} and \textit{Life:die}, probably because the prototype seed triggers are not properly selected. (4) The structure based prompt outperforms the other prompts on \textit{Life:Injure} as \textit{Life:Injure} events require the existence of a person or victim. (5) APEX prompt shows consistently (almost) best performance on all the event types, due to the fact that it combines all the information of other prompts. (6) We also observe that the performance of \textit{Life:Be-Born}, \textit{Life:Die}, \textit{Life:Marry}, and \textit{Personnel:Elect} based on various forms of prompts are consistently better than the other types as the intrinsic semantics of those types the corresponding event triggers are concentrated.

\paragraph{Zero-shot Event Detection} The proposed prompt-based method is more affordable to be generalized comparing with the prior state-of-the-art approach~\cite{zhang-etal-2021-zero}. The average length of created APEX prompts is less than 20 tokens, thus manually creating them won't take much human effort. On the contrary,~\newcite{zhang-etal-2021-zero} requires a large collection of anchor sentences to perform zero-shot event detection, e.g., 4,556,237 anchor sentences for ACE and ERE. This process is time consuming and expensive.

\paragraph{Remaining Challenges} We have demonstrated that a proper description can provide much better performance for both supervised and weakly supervised event detection. However, the event types from most existing ontologies are not properly defined. For example, in ACE annotation guideline~\cite{ldc_ace05}, \textit{transfer-money} is defined as ``\textit{giving, receiving, borrowing, or lending money when it is not in the context of purchasing something}'', however, it's hard for the model to accurately interpret it, especially the constraints ``\textit{not in the context of purchasing something}''. 
In addition, many event types from MAVEN, e.g., \textit{Achieve}, \textit{Award}, and \textit{Incident}, are not associated with any definitions. 
A potential future research direction is to leverage mining-based approaches or state-of-the-art generators to automatically generate a comprehensive event type description based on various sources, such as annotation guidelines, example annotations, and external knowledge bases.

\section{Conclusion}
\label{sec:conclusion}

We investigate a variety of prompts to represent the semantics of event types, and leverage them with a unified framework for supervised, few-shot and zero-shot event detection. Experimental results demonstrate that, a well-defined and comprehensive description of event types can significantly improve the performance of event detection, especially when the annotations are limited (few-shot event detection) or even not available (zero-shot event detection), with up to 24.3\% F-score gain over the prior state of the art. In the future, we will explore mining-based or generation-based approaches to automatically generate a comprehensive description of each event type from available resources and external knowledge base.

\bibliography{ref}
\bibliographystyle{acl_natbib}
\clearpage
\appendix


\section{APEX prompt examples for ACE}
\begin{table*}[!p]
\centering
\small  
\begin{tabular}{L|P}
\toprule
Event Rep Type & Comprehensive Prompt\\
\midrule
Business:Declare-Bankruptcy & Declare Bankruptcy [SEP] bankruptcy bankruptcies bankrupting [SEP] Organization request legal protection from debt collection at a Place \\\midrule
Business:End-Org & End Organization [SEP] dissolving disbanded [SEP] an Organization goes out of business at a Place \\\midrule
Business:Merge-Org & Merge Organization [SEP] merging merger [SEP] two or more Organizations come together to form a new organization at a Place \\\midrule
Business:Start-Org & Start Organization [SEP] founded [SEP] an Agent create a new Organization at a Place \\\midrule
Conflict:Attack & Attack [SEP] invaded airstrikes overthrew ambushed [SEP] An Attacker physically attacks a Target with Instrument at a Place \\\midrule
Conflict:Demonstrate & Demonstrate [SEP] demonstrations protest strikes riots [SEP] Entities come together in a Place to protest or demand official action \\\midrule
Contact:Meet & Meet [SEP] reunited retreats [SEP] two or more Entities come together at same Place and interact in person \\\midrule
Contact:Phone-Write & Phone Write [SEP] emailed letter [SEP] phone or written communication between two or more Entities \\\midrule
Justice:Acquit & Acquit [SEP] acquitted [SEP] a trial of Defendant ends but Adjudicator fails to produce a conviction at a Place \\\midrule
Justice:Appeal & Appeal [SEP] appeal [SEP] the decision for Defendant of a court is taken to a higher court for Adjudicator review with Prosecutor \\\midrule
Justice:Arrest-Jail & Arrest Jail [SEP] arrested locked [SEP] the Agent takes custody of a Person at a Place \\\midrule
Justice:Charge-Indict & Charge Indict [SEP] indictment [SEP] a Defendant is accused of a crime by a Prosecutor for Adjudicator \\\midrule
Justice:Convict & Convict [SEP] pled guilty convicting [SEP] an Defendant found guilty of a crime by Adjudicator at a Place \\\midrule
Justice:Execute & Execute [SEP] death [SEP] the life of a Person is taken by an Agent at a Place \\\midrule
Justice:Extradite & Extradite [SEP] extradition [SEP] a Person is sent by an Agent from Origin to Destination \\\midrule
Justice:Fine & Fine [SEP] payouts financial punishment [SEP] a Adjudicator issues a financial punishment Money to an Entity at a Place \\\midrule
Justice:Pardon & Pardon [SEP] pardoned lift sentence [SEP] an Adjudicator lifts a sentence of Defendant at a Place \\\midrule
Justice:Release-Parole & Release Parole [SEP] parole [SEP] an Entity ends its custody of a Person at a Place \\\midrule
Justice:Sentence & Sentence [SEP] sentenced punishment [SEP] the punishment for the defendant is issued by a state actor \\\midrule
Justice:Sue & Sue [SEP] lawsuits [SEP] Plaintiff initiate a court proceeding to determine the liability of a Defendant judge by Adjudicator at a Place \\\midrule
Justice:Trial-Hearing & Trial Hearing [SEP] trial hearings [SEP] a court proceeding initiated to determine the guilty or innocence of a Person with Prosecutor and Adjudicator at a Place \\\midrule
Life:Be-Born & Be Born [SEP] childbirth [SEP] a Person is born at a Place \\\midrule
Life:Die & Die [SEP] deceased extermination [SEP] life of a Victim ends by an Agent with Instrument at a Place \\
\bottomrule
\end{tabular}
\caption{APEX templates for ACE event types}
\label{tab:ACE prompt}
\end{table*}

\begin{table*}
\centering
\small  

\begin{tabular}{L|P}

\toprule
Event Rep Type & Comprehensive Prompt\\
\midrule
Life:Divorce & Divorce [SEP] people divorce [SEP] two Person are officially divorced at a place \\\midrule
Life:Injure & Injure [SEP] hospitalised paralyzed dismember [SEP] a Victim experiences physical harm from Agent with Instrument at a Place \\\midrule
Life:Marry & Marry [SEP] married marriage marry [SEP] two Person are married at a Place \\\midrule
Movement:Transport & Transport [SEP] arrival travels penetrated expelled [SEP] an Agent moves an Artifact from Origin to Destination with Vehicle at Price \\\midrule
Personnel:Elect & Elect [SEP] reelected elected election [SEP] a candidate Person wins an election by voting Entity at a Place \\\midrule
Personnel:End-Position & End Position [SEP] resigning retired resigned [SEP] a Person stops working for an Entity or change office at a Place \\\midrule
Personnel:Nominate & Nominate [SEP] nominate [SEP] a Person is nominated for a new position by another Agent at a Place \\\midrule
Personnel:Start-Position & Start Position [SEP] hiring rehired recruited [SEP] a Person begins working for an Entity or change office at a Place \\\midrule
Transaction:Transfer-Money & Transfer Money [SEP] donations reimbursing deductions [SEP] transfer Money from the Giver to the Beneficiary or Recipient at a Place \\\midrule
Transaction:Transfer-Ownership & Transfer Ownership [SEP] purchased buy sell loan [SEP] buying selling loaning borrowing giving receiving of Artifacts from Seller to Buyer or Beneficiary at a Place at Price \\
\bottomrule

\end{tabular}

\caption{APEX templates for ACE event types (continued)}
\label{tab:ACE prompt2}
\end{table*}

\end{document}